\documentclass[sigconf]{acmart}

\usepackage{xspace}
\usepackage{times}
\usepackage{epsfig}
\usepackage{graphicx}
\usepackage{amsmath}
\usepackage{booktabs}
\usepackage{multirow}

\usepackage{algorithm}
\usepackage{algpseudocode}
\usepackage{todonotes}
\usepackage{subcaption}

\usepackage{xcolor}

\def\eg{\emph{e.g.}}
\def\ie{\emph{i.e.}}

\setcopyright{acmlicensed}
\copyrightyear{2025}
\acmYear{2025}
\acmDOI{XXXXXXX.XXXXXXX}
\acmConference[ACM MM, 2025]{}{Dublin}{Ireland}
\acmISBN{978-1-4503-XXXX-X/2018/06}

\renewcommand\footnotetextcopyrightpermission[1]{}

\begin{document}

\fancypagestyle{standardpagestyle}{
  \fancyhf{}
  \renewcommand{\headrulewidth}{0pt}
  \renewcommand{\footrulewidth}{0pt}
}
\pagestyle{standardpagestyle} 



\title{SynergyAmodal: Deocclude Anything with Text Control}
\makeatletter
\def\@titlefont{\rmfamily\bfseries\fontsize{20}{24}\selectfont}
\makeatother


\author{\textbf{Xinyang Li\quad Chengjie Yi\quad Jiawei Lai\quad Mingbao Lin \\ Yansong Qu\quad Shengchuan Zhang\quad Liujuan Cao}}

\email{imlixinyang@gmail.com, caoliujuan@xmu.edu.cn}

\affiliation{%
    \institution{Key Laboratory of Multimedia Trusted Perception and Efficient Computing, \\ Ministry of Education of China, Xiamen University, China}
    \country{}
}

\authornote{Corresponding author.}

\begin{abstract}
Image deocclusion (or amodal completion) aims to recover the invisible regions (\ie, shape and appearance) of occluded instances in images. Despite recent advances, the scarcity of high-quality data that balances diversity, plausibility, and fidelity remains a major obstacle.
To address this challenge, we identify three critical elements: leveraging in-the-wild image data for diversity, incorporating human expertise for plausibility, and utilizing generative priors for fidelity.
We propose SynergyAmodal, a novel framework for co-synthesizing in-the-wild amodal datasets with comprehensive shape and appearance annotations, which integrates these elements through a tripartite data-human-model collaboration.
First, we design an occlusion-grounded self-supervised learning algorithm to harness the diversity of in-the-wild image data, fine-tuning an inpainting diffusion model into a partial completion diffusion model.
Second, we establish a co-synthesis pipeline to iteratively filter, refine, select, and annotate the initial deocclusion results of the partial completion diffusion model, ensuring plausibility and fidelity through human expert guidance and prior model constraints.
This pipeline generates a high-quality paired amodal dataset with extensive category and scale diversity, comprising approximately 16K pairs.
Finally, we train a full completion diffusion model on the synthesized dataset, incorporating text prompts as conditioning signals.
Extensive experiments demonstrate the effectiveness of our framework in achieving zero-shot generalization and textual controllability.
Our code, dataset, and models will be made publicly available at {\color{blue}\url{https://github.com/imlixinyang/SynergyAmodal}}.
\end{abstract}

\begin{teaserfigure}
  \includegraphics[width=\textwidth]{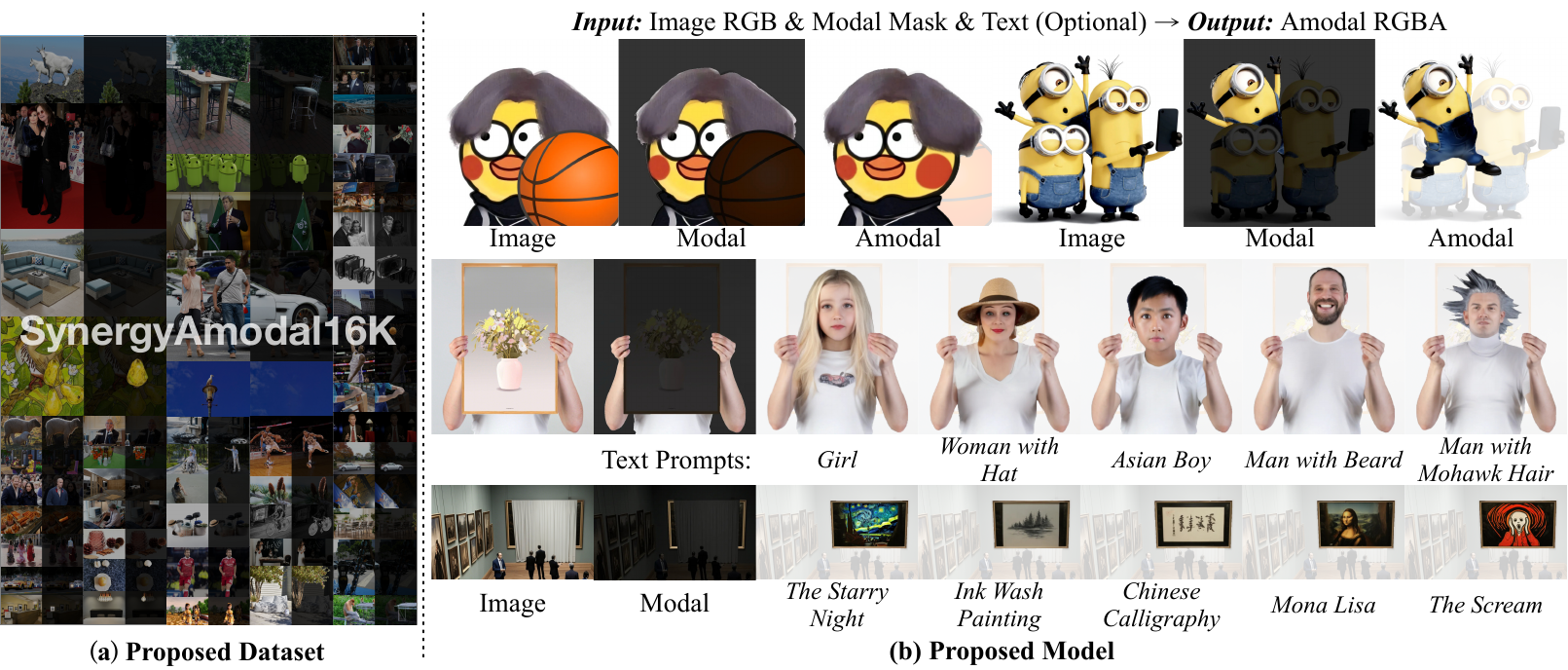}
  \caption{
  In this work, we introduce a dataset and a model tailored for amodal completion.
  \textbf{(a)} SynergyAmodal16K dataset, which covers a wide range of scenes and object categories.
  \textbf{(b)} DeoccAnything model. Given an image and a modal mask, the model can generate the amodal RGBA representation and control the generation process via open-world text prompts.
  }
  \Description{}
  \label{fig:teaser}
\end{teaserfigure}


\maketitle

\begin{figure*}[!h]
    \centering
    \includegraphics[width=1\linewidth]{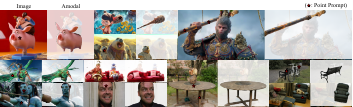}
    \caption{
    \textbf{
    Point-prompted amodal completion by combining SAM~\cite{kirillov2023segment} and the proposed DeoccAnything model.
    } 
    }
    \label{fig.generation}
    \Description{}
\end{figure*}

\section{Introduction}

Recovering complete amodal representations from occluded monocular images is a challenging yet impactful task, with applications spanning AIGC, autonomous driving, and robotics.
A fundamental obstacle of this task lies in the inherent difficulty of acquiring complete amodal ground truth for occluded instances.
Based on the types of relied priors for acquiring amodal data, we broadly categorize existing methods into three classes: \textbf{data}-driven, \textbf{human}-driven, and \textbf{model}-driven.
Data-driven approaches either construct occluded instances from complete instances in existing datasets~\cite{liu2024object} or introduce additional occlusions to train partial deocclusion models~\cite{zhan2020self}. While leveraging in-the-wild datasets enables strong generalization capabilities, the constructed data often lacks physical plausibility, leading to degraded performance under complex occlusions and scale variations.
Human-driven approaches primarily involve direct human annotation of amodal shapes~\cite{zhu2017semantic, qi2019amodal} or indirect acquisition of amodal shapes through human-labeled 3D semantics~\cite{zhan2024amodal}. While human expertise ensures plausible occlusion relationships, these methods are often constrained by limited diversity and predominantly focus on shape-only completion.
Model-driven approaches~\cite{ao2024open, tudosiu2024mulan} utilize priors from pre-trained models (\eg, SAM~\cite{kirillov2023segment}, VLLM~\cite{lai2024lisa}, and Stable Diffusion~\cite{rombach2022high}) for deocclusion tasks. While they can achieve remarkable visual fidelity in certain cases, their lack of domain-specific expertise in deocclusion results in limited success rates.

\begin{figure*}[!t]
    \centering
    \includegraphics[width=1\linewidth]{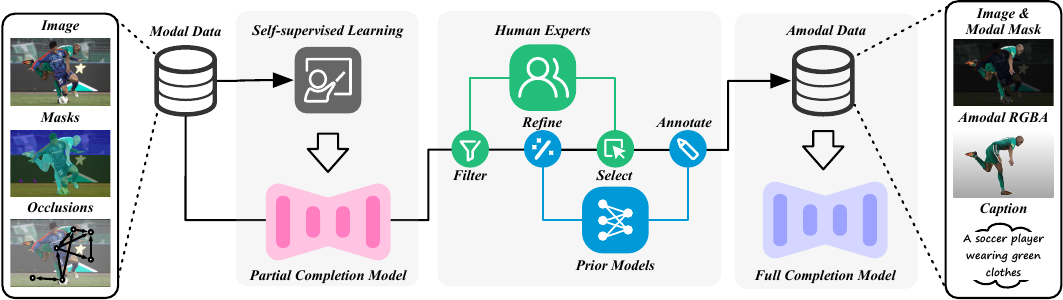}
    \caption{
    \textbf{Framework.}
    We first utilize modal data (\ie, EntitySeg~\cite{qi2022high}) to train a partial completion model using the proposed self-supervised deocclusion learning algorithm.
    Next, we employ a data-human-model collaborative pipeline to generate high-quality amodal data with rich annotations.
    Finally, we train a full completion model using the synthesized amodal data.
    }
    \label{fig.pipeline}
    \Description{}
\end{figure*}

We observe that these methods, by incorporating different types of priors, introduce indispensable strengths: generalization capabilities enabled by diverse data, plausible occlusion reasoning derived from human expertise, and high fidelity provided by strong prior models.
We propose the \textbf{SynergyAmodal} framework to unify the utilization of these priors for constructing novel deocclusion datasets and models as shown in Fig.~\ref{fig.pipeline}. Our contributions include:

$\bullet$ 
A \textbf{data-human-model co-synthesis pipeline} for amodal data generation. 
In this pipeline, we first introduce an occlusion-aware self-supervised deocclusion algorithm capable of leveraging diverse modal data to train a partial completion model. 
This model is then guided by human expertise to filter and select physically plausible results, and further refined and annotated using prior models, yielding high-quality paired deocclusion data efficiently.
%

$\bullet$ 
\textbf{SynergyAmodal16K}, a high-quality amodal dataset of 16K samples based on the EntitySeg~\cite{qi2022high} dataset, which encompasses a wide variety of instance categories and occlusion scenarios, along with high-quality annotations for amodal shape, appearance, and captions.
Some examples are shown in Fig.~\ref{fig:teaser}~(a).

$\bullet$ 
\textbf{DeoccAnything}, an amodal completion diffusion model tuned from an inpainting diffusion model with the proposed dataset, which exhibits generalization capabilities and textual controllability for diverse deocclusion cases, as shown in Fig.~\ref{fig:teaser}~(b). By introducing SAM~\cite{kirillov2023segment} for interactive mask generation, the model can also support point-prompted amodal completion as shown in Fig.~\ref{fig.generation}.

$\bullet$ 
We demonstrate the superiority of the proposed framework over various baseline methods through qualitative and quantitative evaluations on multiple benchmarks and metrics.

\section{Related Works}

\noindent
\textbf{Amodal completion} 
focuses on reconstructing full object shape and appearance based on partially occluded observation.
Early approaches~\cite{kimia2003euler, lin2016computational, silberman2014contour} utilize geometric heuristics (\eg, Euler spirals, Bézier curves or curve primitives) to extrapolate invisible boundaries under predefined occlusion ordering, yet remain restricted to simple shapes.
Subsequent efforts leverage supervised learning with synthetic datasets. However, these methods are constrained to specific object categories, such as vehicles~\cite{ling2020variational, yan2019visualizing, zheng2021visiting}, humans~\cite{zhou2021human, zhang2022face}, and food~\cite{papadopoulos2019make}, while others focus on toy datasets~\cite{burgess2019monet, engelcke2019genesis, greff2019multi} or synthetic indoor objects~\cite{dhamo2019object, ehsani2018segan, zheng2021visiting}.
Recent advances~\cite{zhan2020self, bowen2021oconet, zhan2024amodal, xu2024amodal, ozguroglu2024pix2gestalt} leverage generative models (\eg, GANs~\cite{goodfellow2014generative} and diffusion models~\cite{ho2020denoising}) for amodal completion. 
However, because the utilized datasets of these methods lack specialization or diversity for general deocclusion tasks, a robust framework for open-world occlusion scenarios remains unavailable.
%

\noindent
\textbf{Amodal datasets} 
are critical for advancing research on image deocclusion and amodal completion. Existing methodologies for constructing such datasets can be categorized into three paradigms. 
Human-driven approaches primarily rely on labor-intensive human annotations of amodal masks (\eg, COCOA~\cite{zhu2017semantic} and KINS~\cite{qi2019amodal}), or leverage human-labeled 3D semantics~\cite{zhan2024amodal} and synthetic 3D models~\cite{ehsani2018segan, dhamo2019object, zheng2021visiting, li2023muva}. 
%
Data-driven methods typically create occlusions via manual instance overlays~\cite{follmann2019learning, breitenstein2022amodal, yan2019visualizing, zhou2021human, ao2024amodal, liu2024object, ozguroglu2024pix2gestalt}.
%
Model-driven solutions~\cite{tudosiu2024mulan} establish training-free pipelines to harness pre-trained models for occlusion tasks.
These approaches are constrained by various factors such as diversity, rationality, and success rate. Therefore, there is still an urgent need to collect a high-quality amodal dataset with both shape and appearance annotations.

\noindent
\textbf{Diffusion models} 
have revolutionized image and video generation through their capacity to synthesize photorealistic content.
Initially, DDPM~\cite{ho2020denoising} iteratively denoises data through a Markov chain process, allowing high-quality synthesis of complex distributions. 
Building on this, Stable Diffusion~\cite{rombach2022high} significantly improves computational efficiency through latent-space modeling, which compresses data representations while preserving quality. Further advancements in training and sampling~\cite{song2020denoising, karras2022elucidating} effectively address computational bottlenecks. Additionally, a variety of studies~\cite{ramesh2021zero, brooks2023instructpix2pix, ramesh2022hierarchical, meng2021sdedit, yang2023paint, zhang2023adding, wang2024instancediffusion} provide additional guidance to enable more controlled and customizable generation processes. Beyond image generation, diffusion models have also been successfully applied to diverse tasks, including depth estimation~\cite{ke2024repurposing}, video generation~\cite{blattmann2023stable, menapace2024snap}, multiview synthesis~\cite{liu2023zero} and 3D generation~\cite{poole2022dreamfusion}.
Among these, Pix2Gestalt~\cite{ozguroglu2024pix2gestalt} pioneers the task-specific latent diffusion models for amodal completion, yet inherits the common limitations of non-plausible datasets. 

Our work, instead, introduces a data-human-model co-synthesis pipeline to generate diverse, plausible, and high-fidelity amodal dataset and finetunes latent diffusion models for modeling, achieving state-of-the-art performance for this challenging task.

\section{Method}

\subsection{Problem Formulation and Overview}

Given a RGB image $x$ and an instance mask $m$, the goal of deocclusion is to predict a new RGBA image $g$ that reveals the complete amodal shape and appearance of the target instance while excluding occluders.
In a supervised setting with ground truth amodal data, this task is formulated as:
\begin{equation}\begin{aligned}
    \{x, m\} \xrightarrow{f_\theta} g,
\end{aligned}\end{equation}
where $f_\theta$ represents a full completion model. 

However, acquiring such ground truth data, which includes both shape and appearance amodal annotations, for in-the-wild images remains a significant challenge.
To utilize diverse in-the-wild image data and generate pseudo-labels for amodal ground truth, we first propose a self-supervised approach to learn a partial completion model.
Briefly, let $x_n = (x \circ m)$ and $g_n = x_n \parallel m$ represent the initial RGBA values of the occludee with $n$ occluders, where $\circ$ is element-wise multiplication and $\parallel$ indicates channel concatenation. The step-by-step deocclusion process progressively removes one occluder at each step:
\begin{equation}\begin{aligned}
    g_n \xrightarrow{p_\theta} g_{n-1} \xrightarrow{p_\theta} \dots \xrightarrow{p_\theta} g_1 \xrightarrow{p_\theta} g,
    \label{eq.2}
\end{aligned}\end{equation}
where $p_\theta$ denotes the partial completion model, as detailed in Sec.~\ref{sec.method.1}.
Despite its generalization, the self-supervised model $p_\theta$ faces challenges in handling complex occlusions, leading to unstable predictions. 
To address this, we introduce a data-human-model co-synthesis pipeline in Sec.~\ref{sec.method.2}, refining the behavior of the self-supervised model through minimal human guidance and strong prior models. 
Finally, we train a full completion model $f_\theta$ using the high-quality generated deocclusion dataset, as detailed in Sec.~\ref{sec.method.3}.

\subsection{Self-supervised Partial Completion}
\label{sec.method.1}

\noindent\textbf{Motivation.}
While amodal datasets remain scarce, modal image datasets such as COCO~\cite{lin2014microsoft}, ADE20K~\cite{zhou2019semantic}, EntitySeg~\cite{qi2022high}, and SA-1B~\cite{kirillov2023segment} have emerged as fundamental resources for visual recognition tasks including object detection~\cite{ren2015faster} and instance segmentation~\cite{he2017mask}.
These datasets, though not explicitly designed for amodal completion, inherently contain both unoccluded complete instances and partially occluded variants with diverse occlusion patterns, offering rich implicit supervision for deocclusion learning.
Recent advances~\cite{zhan2020self, ozguroglu2024pix2gestalt} have demonstrated the feasibility of self-supervised learning paradigms that leverage these inherent data properties for deocclusion, inspiring our order-aware extension.

Our key innovation lies in addressing the iterative nature of step-by-step deocclusion through order-grounded supervision.
As formalized in Eq.~\ref{eq.2}, directly learning the mapping from intermediate state $g_n$ to $g_{n-1}$ faces inherent ambiguity.
To resolve this, we propose a learning strategy where the model first learns to remove synthetically added occluders ($g_{n+1} \rightarrow g_{n}$) before tackling natural occlusions ($g_n \rightarrow g_{n-1}$).
While building upon the SSSD framework~\cite{zhan2020self}, we crucially introduce order-aware occlusion synthesis to resolve the dual-occlusion ambiguity that plagues existing approaches (detailed in Appendix). 
This is achieved through seamless integration of occlusion order annotations predicted by InstaOrder~\cite{lee2022instance}.
To align training and inference dynamics, we implement a consistent data construction process and iterative inference process.
Specifically, we define the step-by-step deocclusion as:
\begin{equation}\begin{aligned}
    g_{i-1} = p_\theta (  g_{i},
     m_{\text{deoccluded}, i}, 
    m_{\text{occluder}, i}, 
    x_{\text{background}, i}),
\end{aligned}\end{equation}
where $i$ is the occlusion step, $m_{\text{deoccluded}, i}$ is the processed region of previous occluders but not included in the deoccluded occludee, $m_{\text{occluder}, i}$ is the processing occluder mask, and $x_{\text{background}, i}$ is the background image.

\noindent\textbf{Algorithm.} 
The training and inference workflows are formalized in Algorithm~\ref{alg1} and Algorithm~\ref{alg2}, respectively.
Specifically, during training, given the image $x$, an instance mask $m_n$, its occluder masks $\{m_{\text{occluder},i}\}_{i=1}^n$, and a randomly generated mask $m_{\text{generated}}$ sampled from other images, we first compute the new occluder mask $m_{\text{occluder},n+1}$ by excluding all existed occluder masks from the generated mask. 
The instance mask $m_{n+1}$ is then updated by removing the new occluder regions. 
The deoccluded mask $m_{\text{deoccluded},n+1}$ is synthesized by randomly sampling from the visible non-occludee areas while excluding the new occluder, where $v_j \sim \text{Bernoulli}(0.5)$ is a random binary value.
The background image $x_{\text{background}, n+1}$ is preserved by masking the original image with the mask which excludes the instance, new occluder and deoccluded mask regions. 
For inference, we adopt an iterative approach working backwards from step $n$ to $1$. 
At each step $i$, we maintain the current generated image $g_i$ and progressively reconstruct the image by predicting previous states using learned model $p_\theta$. 
The background image $x_{\text{background}, i}$ is computed similarly to the training process, ensuring consistency between the training and inference processes. 
The deoccluded mask $m_{\text{deoccluded}, i}$ is continuously updated to track newly revealed regions as occlusions are removed. 
This iterative process continues until we obtain the final reconstructed image $g=g_0$, which represents our initial completely deoccluded result.

To enhance the model's parallel processing capability for complex occlusion scenarios, we further extend the algorithm through adding multiple occluders at once during training. 
This extension enables simultaneous multi-occluder removal through explicit exposure the model for combinatorial occlusion patterns, which improves computational efficiency by reducing required inference steps and mitigates possible error propagation risks.

\begin{algorithm}[t]
    \caption{Training Process}
    \label{alg1}
    \leftline{\hspace*{0.02in} {\bf Input:} %
    $x,m_n,\{m_{\text{occluder},i}\}_{i=1}^n,m_{\text{generated}}$}
    \leftline{\hspace*{0.02in} {\bf Output:} %
    $g_{n+1},m_{\text{occluder},n+1},m_{\text{deoccluded},n+1},x_{\text{background}, n+1}$}
    \begin{algorithmic}[1]
    \State $m_{\text{occluder},n+1} \leftarrow m_{\text{generated}} \setminus \bigcup_i m_{\text{occluder},i}$
    \State $m_{n+1} \leftarrow m_n \setminus m_{\text{occluder},n+1}$
    \State $m_{\text{deoccluded},n+1} \leftarrow \bigcup_j (v_j \cdot m_{\text{non-occludee}, j} ) \setminus m_{\text{occluder},n+1}$
    \State $x_{\text{background}, n+1} \leftarrow x \circ (\overline{m}_{\text{deoccluded}, n+1} \cap \overline{m}_{\text{occluder}, n+1} \cap \overline{m}_{n+1}) $
    \State \Return $g_{n+1},m_{\text{occluder}, n+1},m_{\text{deoccluded},n+1},x_{\text{background}, n+1}$
    \end{algorithmic}
\end{algorithm}

\begin{algorithm}[t]
    \caption{Inference Process}
    \label{alg2}
    \leftline{\hspace*{0.02in} {\bf Input:} %
    $x$, $m_n$, $\{m_{\text{occluder},i}\}_{i=1}^n$}
    \leftline{\hspace*{0.02in} {\bf Output:} %
    $g$}
    \begin{algorithmic}[1]
    \State $g_n \leftarrow (x \circ m_n) \parallel m_n$
    \State $m_{\text{deoccluded}, n} \leftarrow \mathbf{0}$
    \For{ $i = n, n-1, \dots, 2, 1$ }
        \State $x_{\text{background}, i} \leftarrow x \circ (\overline{m}_{\text{deoccluded}, i} \cap \overline{m}_{\text{occluder}, i} \cap \overline{m}_i) $
        \State $g_{i-1} \leftarrow p_\theta(g_i, m_{\text{deoccluded}, i}, m_{\text{occluder}, i}, x_{\text{background},i})$
        \State $m_{\text{deoccluded},i-1} \leftarrow m_{\text{deoccluded},i} \cup (m_{\text{occluder},i} \setminus m_i)$
    \EndFor
    \State $g \leftarrow g_0$
    \State \Return $g$
    \end{algorithmic}
\end{algorithm}

\noindent\textbf{Modeling.} 
We implement the partial completion function through a repurposed latent diffusion model architecture. Building upon the pre-trained Stable Diffusion 2 Inpainting model~\cite{rombach2022high}, we strategically modify its conditioning mechanism while preserving essential generalization capabilities.   
The original denoising une t of Stable Diffusion 2 Inpainting model accepts three conditions: a latent encoding of a masked image; a downsampled inpainting mask; and a text prompt.
To adapt it for self-supervised deocclusion training, we need to adjust the conditions.
To maintain generalization ability and make the adaption smooth, we maintain the original functionality of the original two conditions as $E(x_{n+1})$ and $m_{\text{occluder}, n+1} \downarrow$ and substitute text prompts with empty embeddings $\phi$ to disable textual conditioning.
Other conditions (\ie, the background image $x_{\text{background}, n+1}$ and the triple masks $ (m_{\text{deoccluded}, n+1}, m_{\text{occluder}, n+1}, m_{n+1})$) are input into the denoising network with a zero-initialized convolutional layer.
Formally, The denoising unet $\epsilon_{p_\theta}$ for partial completion is optimized through the following noise prediction objective:
\begin{equation}\begin{aligned}
    & \mathcal{L}_{p_\theta}=\|\epsilon-
    \epsilon_{p_\theta}(z_t, t, \phi, m_{\text{occluder}, n+1} \downarrow,
    \mathcal{E}(x_{n+1}),
    \\
    & \mathcal{E}(x_{\text{background}}), \mathcal{E}(m_{\text{deoccluded}, n+1} \parallel  m_{\text{occluder}, n+1} \parallel m_{n+1}))\|^2_2,
\end{aligned}\end{equation}
where the noisy latent is derived by 
$
    z_t = \sqrt{\alpha_t} \mathcal{E}(x_{n}) + \sqrt{1-\alpha_t} \epsilon
$
, $\alpha_t$ is a monotonically decreasing noise schedule, $t$ is the sampled timestep, and $\epsilon \sim \mathcal{N} (0, 1)$ is a random noise.
During inference, given the occluded instance image $g_{i}$ and other conditions, the model can progressively predict the latent code of the partially deoccluded instance image $x_{i-1}$ by iteratively latent denoising and decoding.
Additionally, to obtain the instance mask $m_i$ efficiently and accurately, we train an extra mask decoder to output mask based on the latent of the instance image, rather than training individual model~\cite{zhan2020self} or using thresholding~\cite{ozguroglu2024pix2gestalt}.

\subsection{Pesudo Data Co-Synthesis}
\label{sec.method.2}

Although the self-supervised model demonstrates fundamental deocclusion capabilities with reasonable generalization, its performance and efficiency remain constrained when handling complex multi-layer occlusion patterns. Moreover, the reliance on pre-annotated occlusion relationships limits its practical applicability.
To develop a production-ready amodal completion model while enhancing deocclusion performance, we propose to generate high-quality deocclusion data through synergistic integration of human expertise and prior model knowledge.
The synthesis process unfolds as follows: We first engage several human annotators to filter instances with diverse occlusion relationships and correct automatically annotated occlusion orders based on their domain knowledge. The filtered instances are either directly marked as \textit{unoccluded} or processed through the self-supervised model to obtain initial deocclusion results. When unsatisfactory outcomes persist after multiple retries, human experts flag these instances as \textit{failed}.
Then, the initial deocclusion results are refined by Stable Diffusion 3~\cite{esser2024scaling} for fidelity and better preservation of original unoccluded regions. 
This refinement phase automatically produces multiple variants by applying different random seeds and noise strengths $(0.5, 0.75, 1)$ to the initial results. Human experts subsequently select the optimal refined outputs through visual inspection.
Finally, the chosen results receive comprehensive annotations through ZIM~\cite{kim2024zim} and InternVL~\cite{chen2024internvl} to automatically generate supplementary information including fine-grained masks and descriptive captions.
This pipeline ultimately yields 16K high-quality deocclusion pairs featuring diverse occlusion scenarios and rich annotations.

\begin{figure}[!t]
    \centering
    \includegraphics[width=1\linewidth]{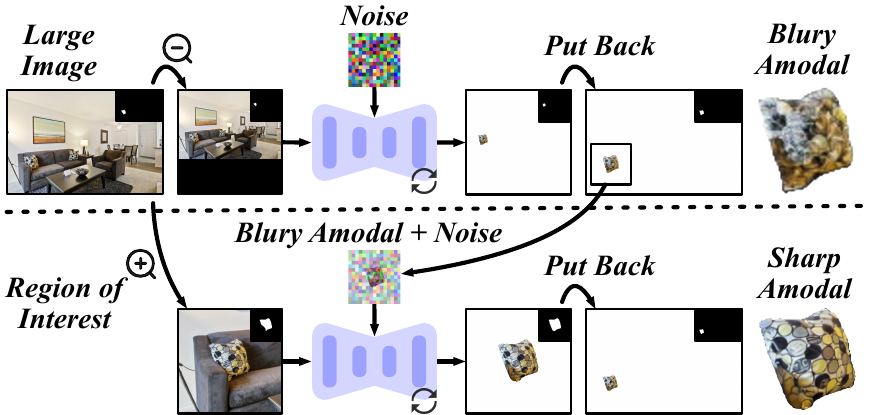}
    \caption{
    \textbf{Global-to-local inference strategy}. 
    }
    \Description{}
    \label{fig.inference}
\end{figure}

\subsection{Pseudo-supervised Full Completion}
\label{sec.method.3}

Building upon the pseudo amodal ground truth acquired through our co-synthesis pipeline, we implement direct supervision for training the full completion model. 
The architectural design maintains consistency with the self-supervised partial completion model with minor changes.
We preserve the original conditions of Stable Diffusion 2 Inpainting model as $\mathcal{E}(x \cdot m)$ and $\overline{m} \downarrow$, and incorporate an additional conditioning signal $\mathcal{E}(x)$ into the denoising unet, enabling stable training convergence and well-maintained generalizable priors.
Furthermore, we reintroduce the text prompt $y$ back into the conditions, leveraging the amodal captions obtained in previous stages.
The training objective of the full completion diffusion model is formally defined as:
\begin{equation}\begin{aligned}
    \mathcal{L}_{f_\theta}=\|\epsilon-
    \epsilon_{f_\theta}(z_t, t, y, \overline{m} \downarrow,
    \mathcal{E}(x \cdot m), \mathcal{E}(x))\|^2_2,
\end{aligned}\end{equation}
where the noisy latent $z_t$ is constructed as
$
    z_t = \sqrt{\alpha_t} \mathcal{E}(x_g) + \sqrt{1-\alpha_t} \epsilon
$ and $x_g$ denotes the RGB channels of the amodal RGBA image $g$.
During inference, the alpha-channel amodal mask is also obtained by decoding the latent of predicted $x_g$ with an extra mask decoder.
See the detailed architecture in our appendix.

Although the size of our synthesized dataset is relatively small, the model still demonstrates excellent generalization performance, benefiting from the diversity of the data, the richness of annotations, and the powerful priors of the pretrained inpainting diffusion model.

\noindent\textbf{Global-to-local inference.}
Given the resolution constraints and information loss inherent in VAE-based latent space, we devise a two-stage inference strategy to enhance visual fidelity when inputting images with larger resolution of the model configurations.
As illustrated in Fig.~\ref{fig.inference}, we first process the entire image (with resizing and padding for non-square inputs) to obtain initial blurry amodal results.
Then, we extract region-of-interest crops guided by the predicted amodal masks, then perform another denoising process with reduced noise strength, utilizing the blurry results as initialization.
This hierarchical process mitigates artifacts caused by the latent space while preserving global contextual understanding, leading to sharper amodal results.

\section{Experiments}

\subsection{Settings}

\noindent
\textbf{Implementation details.}
We utilize the EntitySeg~\cite{qi2022high} as the modal dataset for its pixel-accurate and category-agnostic segmentation annotation to construct SynergyAmodal16K dataset. 
The data co-synthesis process is conducted by three expert annotators, spending about 200 hours in total.
For training, both the partial and full completion diffusion models undergo 50K optimization steps with batch size 64, consuming 4 days on 4×A100 40G GPUs.
We maintain the identical spatial resolutions to Stable Diffusion 2 (\ie, pixel space: 512×512, latent space: 64×64).
We employ 50-step DDIM sampling~\cite{song2020denoising} during evaluation.
%

\noindent
\textbf{Baselines.}
We conduct comprehensive comparisons with state-of-the-art approaches across different paradigms:
\textbf{SSSD}~\cite{zhan2020self} is a data-driven self-supervised deocclusion method that employs two separate models: one for generating amodal masks and another for producing corresponding RGB content using a unet-based regression model and generative adversarial networks (GANs), respectively.
\textbf{SDAmodal}~\cite{zhan2024amodal} is a human-driven supervised deocclusion method that leverages both synthesized and human-annotated datasets for training.
It also proposes utilizing the intermediate feature maps of Stable Diffusion to enhance the generalization of the deocclusion model.
However, since SDAmodal can only generate amodal masks, we use the same inpainting model (\ie, Stable Diffusion 2 Inpainting) to generate RGB content within the amodal mask.
\textbf{StableDiffusion}. To provide a baseline for model-driven methods, we implement a model by combining a Vision-Language Large Model (InternVL~\cite{chen2024internvl}), an inpainting model (Stable Diffusion 2 Inpainting), and a segmentation model (ZIM~\cite{kim2024zim}).
\textbf{Pix2Gestalt}~\cite{ozguroglu2024pix2gestalt} is another data-driven self-supervised deocclusion method that adopts a different data construction process and a diffusion-based architecture.
It constructs occlusion data using unoccluded instances in real-world images. However, this approach is limited to single-instance occlusion and suffers from poor physical plausibility of data.


\subsection{Quantitative Comparison}

\begin{figure}[!t]
    \centering
    \includegraphics[width=1\linewidth]{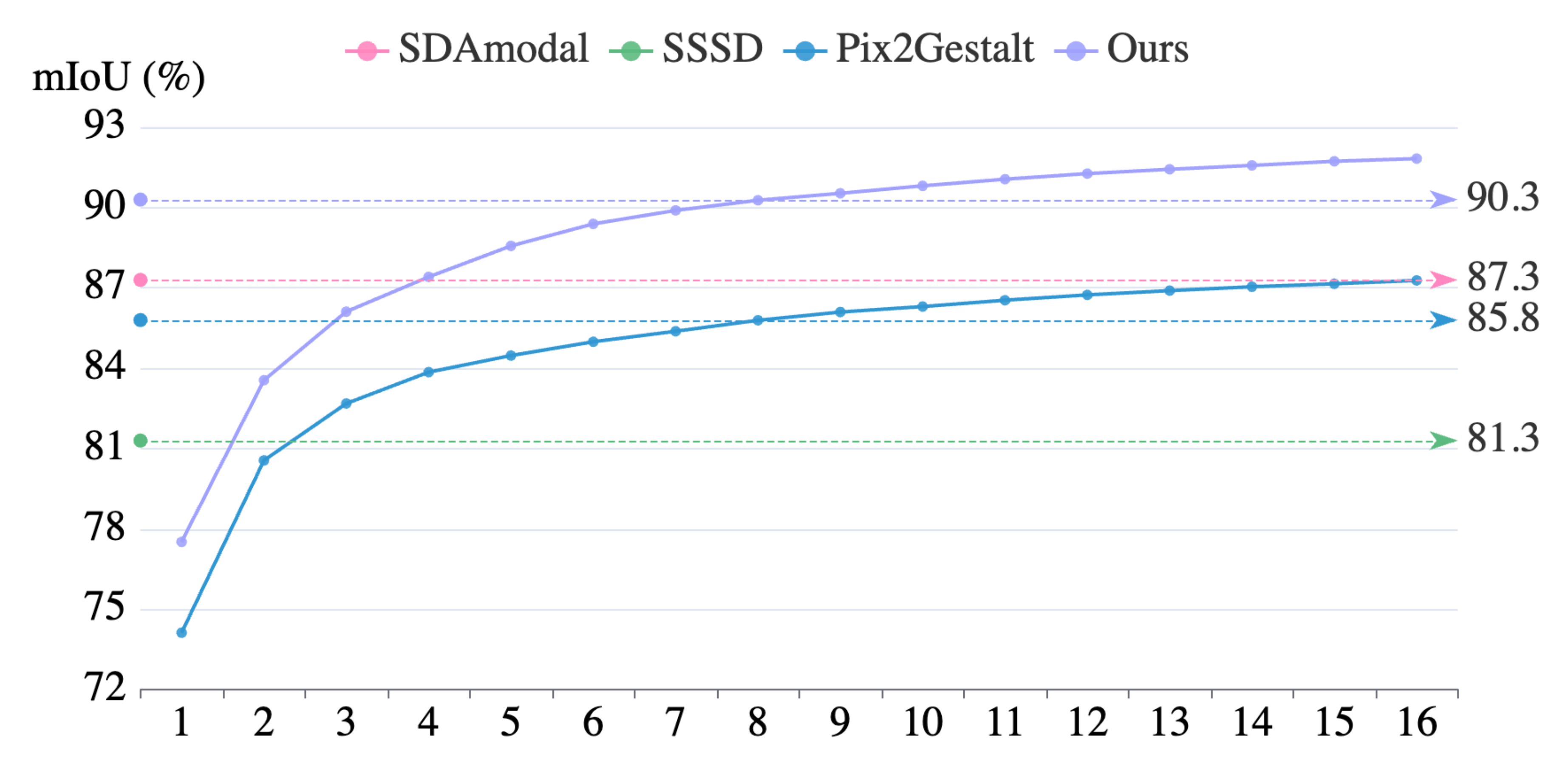}
    \caption{
    \textbf{Effect of different variation number.}
    }
    \Description{}
    \label{fig.quantitative-1}
\end{figure}

We conduct a quantitative comparison between our framework and baseline models, with the results presented in Tab.~\ref{tab.quantitative}.
The evaluation is performed on two widely-adopted benchmarks: COCOA~\cite{zhu2017semantic} and BSDSA~\cite{martin2001database}.
Both datasets consist of real-world images accompanied by expert-annotated amodal instance masks.
Specifically, The evaluation set of COCOA contains 12,753 amodal annotations across 1323 images, while BSDSA includes 4175 annotated objects distributed among 100 images.
To assess the amodal segmentation performance, we employ the mean Intersection-over-Union (mIoU) metric, which measures the overlap between the predicted amodal instance masks and the ground truth masks.
For generative methods (\ie, Pix2Gestalt and our approach), which may produce varying deocclusion results due to random noise, we generate 8 samples for each instance and select the best-performing variation for evaluation.
Our method achieves the highest mIoU scores on both datasets.
To provide a more detailed comparison, we further analyze the mIoU performance of competitive methods with varying numbers of generated variations, as illustrated in Fig.~\ref{fig.quantitative-1}.
The results demonstrate that our method outperforms the state-of-the-art self-supervised method SSSD and supervised method SDAmodal with only 2 and 4 variations on COCOA dataset, respectively, noted that SDAmodal is directly trained on the COCOA dataset.

\begin{table}[!t]
    \centering
    \scalebox{0.93}{%
    \begin{tabular}{lcccc}
    \toprule
\multirow{2}{*}{Method}                         & \multicolumn{2}{c}{COCOA} & \multicolumn{2}{c}{BSDSA} \\
    & mIoU(\%)$\uparrow$         & FID$\downarrow$         & mIoU(\%)$\uparrow$         & FID$\downarrow$         \\
\midrule
SDAmodal~\cite{zhan2024amodal}&87.3&11.0&89.7&37.1\\
\midrule
SSSD~\cite{zhan2020self}&81.3&12.8&34.1&92.9\\
StableDiffusion~\cite{rombach2022high}&58.5&15.1&63.8&46.4\\
Pix2Gestalt~\cite{ozguroglu2024pix2gestalt}&85.8&13.6&88.3&39.8\\
\midrule
\textbf{Ours}&\textbf{90.3}&\textbf{9.5}&\textbf{90.2}&\textbf{34.3}\\
    \bottomrule
    \end{tabular}%
    }   
    \vspace{4pt}
    \caption{Quantitative comparison on COCOA and BSDSA.}
    \Description{}
    \label{tab.quantitative}
\end{table}

\begin{figure}[!t]
    \centering
    \includegraphics[width=1\linewidth]{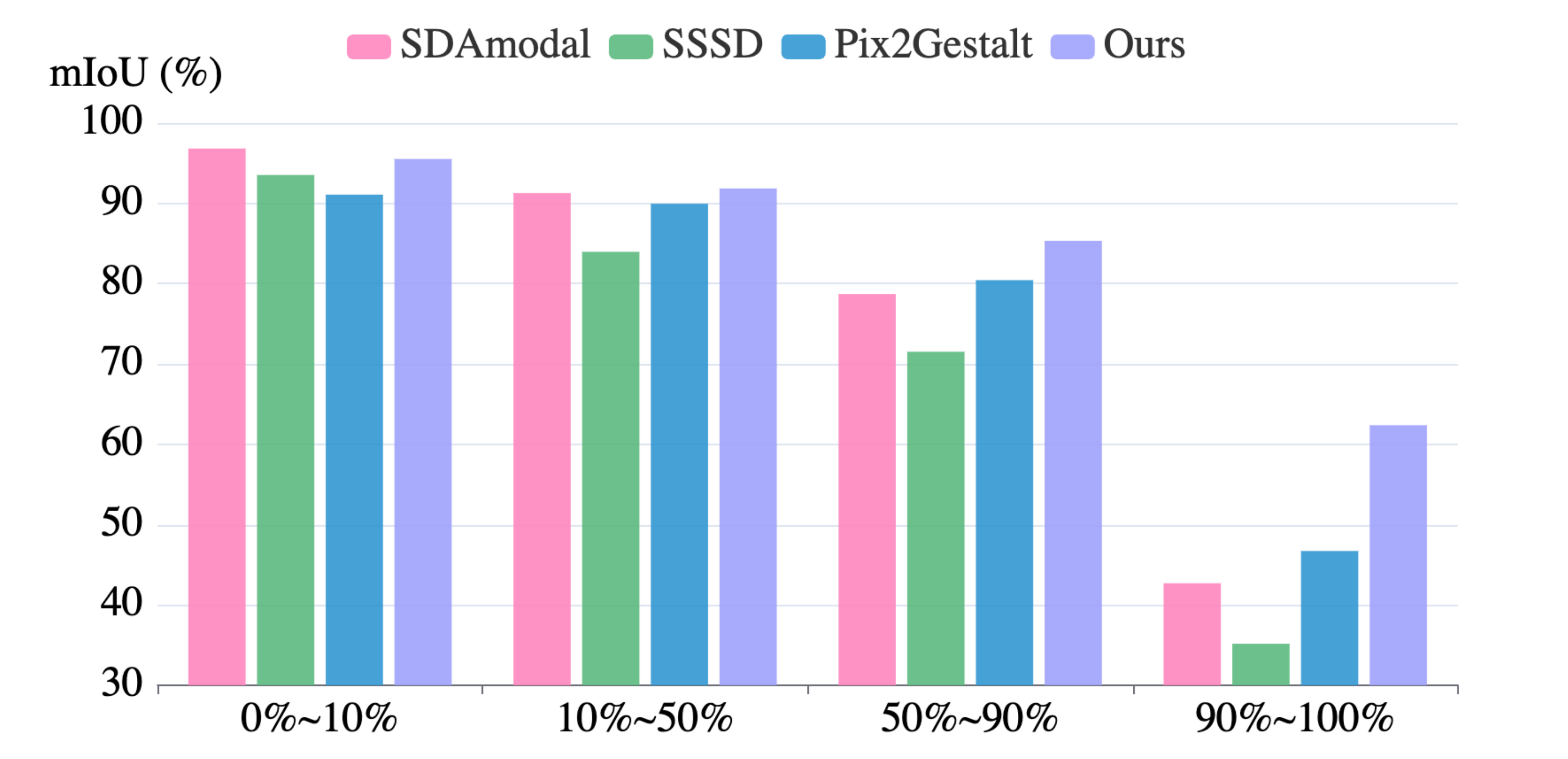}
    \caption{
    \textbf{Effect of different occlusion percentages.}
    }
    \Description{}
    \label{fig.quantitative-2}
\end{figure}

In order to comprehensively observe the model's deocclusion performance under different occlusion percentages, we plot a bar chart in Fig.~\ref{fig.quantitative-2}. It can be observed that:
SDAmodal, trained directly on COCOA’s amodal annotations, achieves the highest mIoU in the 0\%\textasciitilde10\% range, demonstrating its effectiveness in minimally occluded scenarios. 
However, as occlusion increases, generative methods like Pix2Gestalt and our model gradually outperform other approaches.
Notably, our model consistently delivers robust performance across all occlusion levels, achieving the highest mIoU in the 10\%\textasciitilde50\%, 50\%\textasciitilde90\%, and 90\%\textasciitilde 100\% ranges and maintaining competitive results in the 0\%\textasciitilde10\% interval. 
This highlights the superior generalization capability of our method in handling diverse occlusion scenarios, validating its effectiveness in real-world amodal completion applications.

To evaluate the fidelity of the deocclusion results across different methods, we employ the Fréchet Inception Distance (FID)~\cite{heusel2017gans} as evaluation metric.
This metric measures the feature similarity between the distribution of real unoccluded instances in the dataset and the generated deoccluded instances of occluded instances.
SSSD demonstrates satisfactory FID scores on the COCOA validation set due to its training on COCOA images. However, its performance deteriorates significantly when tested on the domain-shifted BSDSA dataset, revealing limited cross-domain generalization capabilities.
SDAmodal achieves the second-best results across both datasets, benefiting from its powerful generative architecture and competent amodal mask prediction capability.
StableDiffusion exhibits suboptimal performance on both benchmarks, primarily attributed to its lack of explicit occlusion reasoning mechanisms required for amodal completion tasks.
Pix2Gestalt shows relatively balanced performance between datasets, indicating moderate generalization potential. Nevertheless, its FID metrics remain constrained by limited resolution and generation quality.
The results demonstrate that our method achieves the best FID scores among all methods for both datasets.
This indicates that the distribution of our deocclusion results is the most consistent with the in-the-wild completed instances, highlighting the superior quality and realism of our model.

\subsection{Qualitative Comparison}

\begin{figure*}[!t]
    \centering
    \includegraphics[width=1\linewidth]{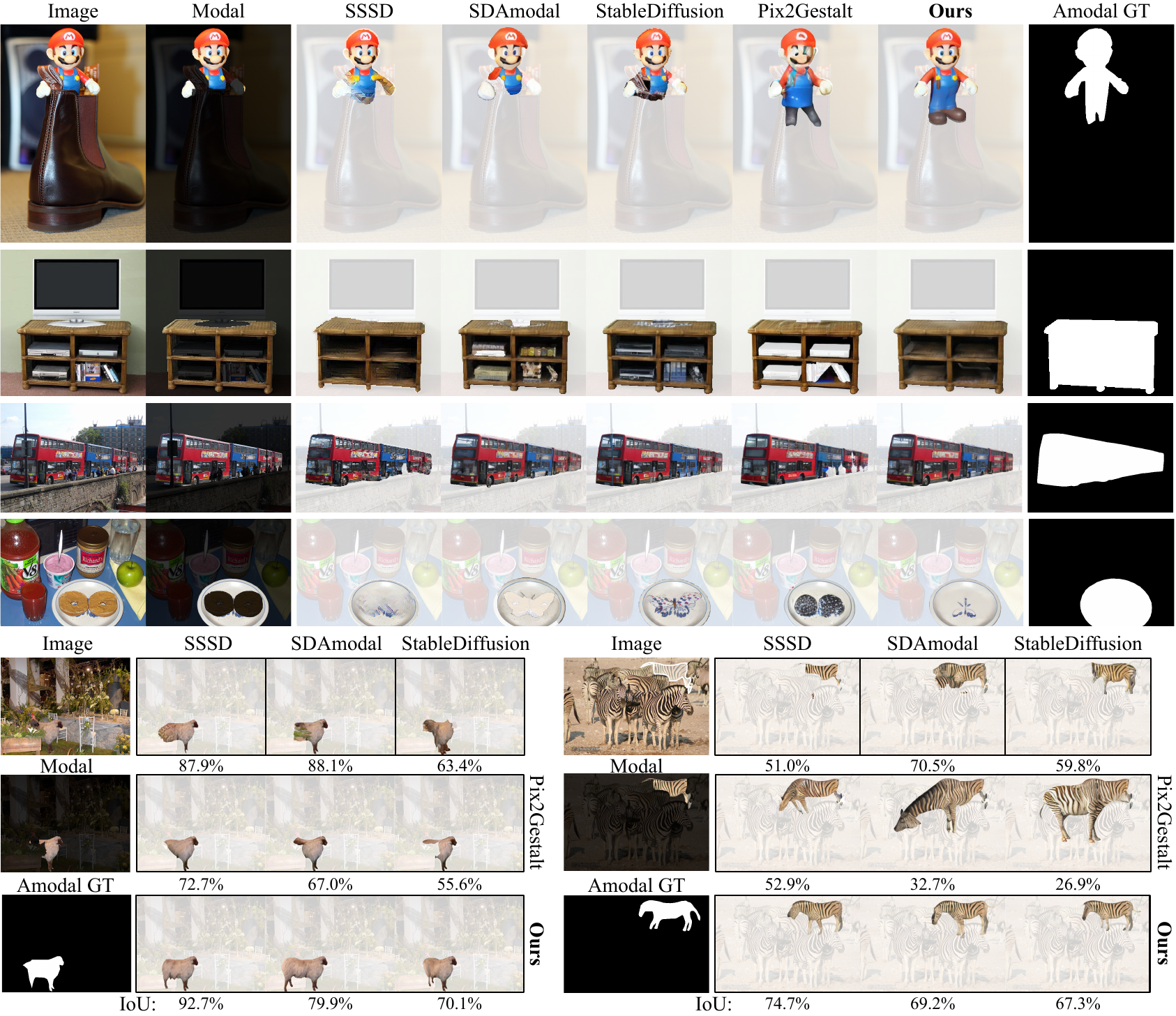}
    \caption{
    Qualitative comparison on COCOA.
    }
    \Description{}
    \label{fig.qualitative}
\end{figure*}

We provide a qualitative comparison between our framework and baseline models, as illustrated in Fig.~\ref{fig.qualitative}.
SSSD demonstrates reasonable performance on simple cases but tends to generate unrealistic shapes and external artifacts in more complex and articulated scenarios.
Due to the lack of specialized deocclusion expertise, the generation results of StableDiffusion often exhibit incomplete or redundant instance contents, resulting in poor performance across most cases.
The outputs of Pix2Gestalt often suffer from blurred original content.
Furthermore, since the data construction process of Pix2Gestalt fails to fully encompass the distribution of real-world deocclusion scenarios, it struggles to achieve satisfactory results in cases involving multiple occluders (\eg, \textit{cabinet} in Row 2 and \textit{bus} in Row 3).
Our method consistently outperforms the baselines in most cases, delivering accurate shapes, high-quality appearances, and effective preservation of the original content.
These results underscore the effectiveness of our data synthesis and training pipeline.

At the bottom of the figure, we also provide a qualitative experiment demonstrating the effect of different variations generated by the generative model, with the IoU metrics for each result labeled below the corresponding amodal images. 
As can be observed, regression-based methods (\ie, SSSD and SDAmodal) tend to produce results resembling an ``average'' outcome. 
While they often achieve decent IoU scores, the actual shapes do not meet the requirements of the deocclusion task. 
Pix2Gestalt still underperforms in cases of complex occlusion, especially when the occlusion rate is high. 
In contrast, our method, despite fluctuations in IoU across different variations, consistently produces realistic shapes and appearances that align with the semantic meaning of the instances. 
This experiment not only highlights the necessity of calculating the mIoU across different variations but also demonstrates the outstanding generalization capability and realism of our method across diverse scenarios.


\subsection{Text Control Results}

\begin{figure*}[!t]
    \centering
    \includegraphics[width=1\linewidth]{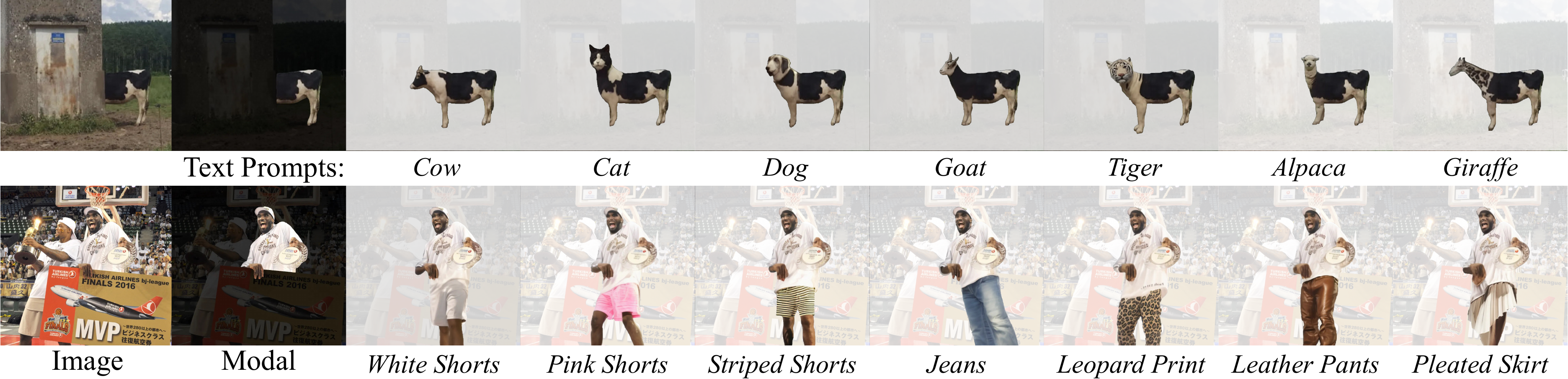}
    \caption{
        \textbf{Results of amodal completion with text control.}
    }
    \label{fig.control}
\end{figure*}

Since our full completion model incorporates text as a conditional input, we can guide the deocclusion generation process by employing different text prompts.
We illustrate this capability with two extra examples in Fig.~\ref{fig.control}.
The results show that the deocclusion results can be well-aligned with the specified textual guidance.
For instance, we can control the species of animals when their head regions are occluded or dictate the clothing style for the obscured lower body of a person.
Even when the prompts describe scenarios that are uncommon in real-world cases, the model still produces reasonable and visually harmonious deocclusion results.
This textual controllability, which is unattainable with traditional amodal completion methods, introduces significant flexibility and creative potential to this task.



\subsection{Ablation Study}

\begin{figure}[!t]
    \centering
    \includegraphics[width=1\linewidth]{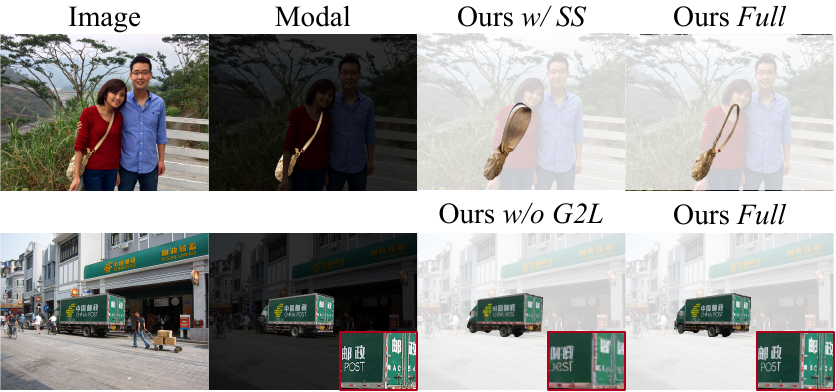}
    \caption{
        \textbf{Qualitative ablation study.}
    }
    \label{fig.ablate}
\end{figure}

To validate the significance of each component in our method, we focus on two key ablations: our self-supervised model (Ours \textit{w/ SS}) and our full completion model without the global-to-local inference strategy (Ours \textit{w/o G2L}).
We evaluate these ablations using the mIoU and FID metrics on both the COCOA and BSDSA datasets.
Ours \textit{w/ SS} achieves slightly worse mIoU and FID scores compared to our full method but performs marginally better in FID compared to Ours \textit{w/o G2L}.
\begin{table}[!t]
    \centering
    \scalebox{0.96}{%
    \begin{tabular}{lcccc}
    \toprule
\multirow{2}{*}{Method}                         & \multicolumn{2}{c}{COCOA} & \multicolumn{2}{c}{BSDSA} \\
    & mIoU(\%)$\uparrow$         & FID$\downarrow$         & mIoU(\%)$\uparrow$         & FID$\downarrow$         \\
\midrule
Ours \textit{w/ SS}&89.9&10.5&88.2&36.1\\
Ours \textit{w/o G2L}&90.1&10.9&90.1&37.4\\
\midrule
Ours \textit{Full}&\textbf{90.3}&\textbf{9.5}&\textbf{90.2}&\textbf{34.3}\\
    \bottomrule
    \end{tabular}%
    }   
    \vspace{10pt}
    \caption{
        \textbf{Quantitative ablation study.}
    }
    \label{tab.1}
\end{table}
\begin{figure}[!t]
    \centering
    \includegraphics[width=1\linewidth]{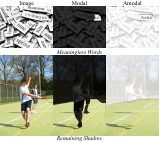}
    \caption{
        \textbf{Failure cases of our model.}
    }
    \label{fig.failure}
\end{figure}
This can be attributed to the self-supervised model's reliance on occlusion annotations, which enables it to scale images into more localized regions during inference.
In contrast, our full completion model requires inputting the original-size image first, which results in blurry outputs.
This blurriness has a minor impact on the FID.
To visualize this phenomenon, we provide a qualitative ablation study as shown in Fig.~\ref{fig.ablate}.
It also demonstrates that self-supervised models are highly prone to over-deocclusion when dealing with finer objects.
Overall, our full completion model equipped with the global-to-local inference strategy achieves the best performance across all metrics, without requirement of occlusion annotations.
%

\subsection{Failure Cases and Limitations }

Although our method can successfully generalize to arbitrary images and instances, there are still cases where deocclusion may fail. We present some of these cases in Fig.~\ref{fig.failure}.
First, for text content, the generated results may include meaningless symbols. This primarily stems from the pretrained model's inherent limitations in handling such content.
Second, the model cannot automatically handle shadows or reflections of other objects caused by lighting, which may result in unwanted information being present in the generated amodal RGB. This could potentially be solved by integrating an additional shadow removal model~\cite{guo2023shadowdiffusion}.
%

\section{Conclusion}

In this work, we present SynergyAmodal, a novel framework for generating high-quality amodal completion datasets and models through a collaborative approach involving in-the-wild image data, human expertise, and strong model priors. 
By integrating in-the-wild image data for diversity, human expertise for plausibility, and model priors for fidelity, we successfully create a diverse and high-fidelity amodal dataset with rich annotations. 
Using this dataset, we train a diffusion model that demonstrates strong performance in terms of zero-shot generalization and textual controllability for open-world amodal completion. 
Our framework effectively addresses the data scarcity challenge in image deocclusion and achieved state-of-the-art results in amodal completion.
Our future works include expanding the dataset scale by introducing human-in-the-loop operations~\cite{li2025rorem, mosqueira2023human}, leveraging more advanced pre-trained image diffusion models~\cite{esser2403scaling, flux2024}, and exploring the potential of our framework in 3D~\cite{wu2025amodal3ramodal3dreconstruction} and video~\cite{lu2025tacotamingdiffusioninthewild} amodal tasks.


\bibliographystyle{ACM-Reference-Format}
\bibliography{sample-base}


\clearpage

\appendix

\section{Dual-Occlusion Ambiguity in SSSD~\cite{zhan2020self}}

\begin{figure}[!h]
    \centering
    \includegraphics[width=1\linewidth]{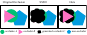}
    \caption{
        \textbf{Dual-occlusion ambiguity and our solution.}
    }
    \label{fig.dual}
\end{figure}

\begin{figure}[!h]
    \centering
    \includegraphics[width=1\linewidth]{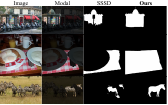}
    \caption{
        \textbf{Qualitative comparison of SSSD and our method on amodal segmentation task.}
    }
    \label{fig.sssd}
\end{figure}

\noindent
We provide a detailed explanation of the dual-occlusion ambiguity of SSSD~\cite{zhan2020self} in Fig.\ref{fig.dual}.
Specifically, when a randomly generated occluder mask \textit{C} obscures another occluder \textit{B} that already partially occludes the occludee \textit{A}, the model incorrectly learns the complete amodal representation as the partially occluded \textit{A}.
By introducing the order-grounded self-supervised deocclusion learning algorithm proposed in this paper, the generated occluder mask \textit{C} is correspondingly adjusted to exclude the partially occluded region of occluder \textit{B}. 
This ensures that, after removing mask \textit{C}, the recovered representation of occluder \textit{B} still reflects the partially occluded \textit{A}, but the model explicitly recognizes that part of \textit{A} remains occluded by \textit{B}.
To better illustrate this ambiguity, we present a scenario where \textit{C} completely occludes \textit{B}. However, this ambiguity persists even when \textit{C} partially occludes \textit{B}, as the model struggles to focus on the semantic information of \textit{B} to perform deocclusion for \textit{A}. This misleads the model into believing that the complete shape of \textit{A} is different from its actual form.
We showcases more amodal segmentation results of SSSD and our method in Fig.~\ref{fig.sssd}.
It can be observed that the dual-occlusion ambiguity severely impacts the success rate of SSSD, whereas our method effectively circumvents this limitation.

\section{Architecture Details of DeoccAnything}


\begin{figure*}[!t]
    \centering
    \includegraphics[width=1\linewidth]{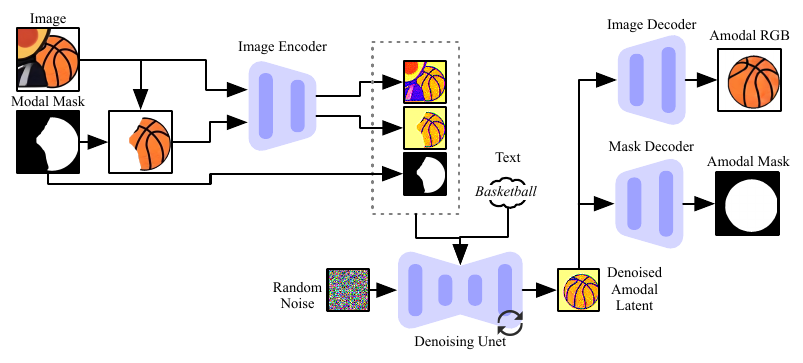}
    \caption{
        \textbf{Detailed architecture of DeoccAnything model.}
    }
    \label{fig.archi}
\end{figure*}

We present the detailed architecture of our DeoccAnything model in Fig.~\ref{fig.archi}.
Specifically, we retain the original conditions of the Stable Diffusion 2 Inpainting model, including: the encoded visible region $\mathcal{E}(x \cdot m)$; (2) the downsampled mask to be completed $\overline{m} \downarrow$; and the text prompt.
An encoded image latent $\mathcal{E}(x)$ is introduced as an extra condition to the denoising unet.
During inference, a randomly sampled noise is denoised through the denoising unet to produce the denoised amodal latent.
The latent is then decoded into the amodal RGB and mask using separate RGB and mask decoders.

\section{Amodal 3D Reconstruction}

\begin{figure*}[!t]
    \centering
    \includegraphics[width=0.87\linewidth]{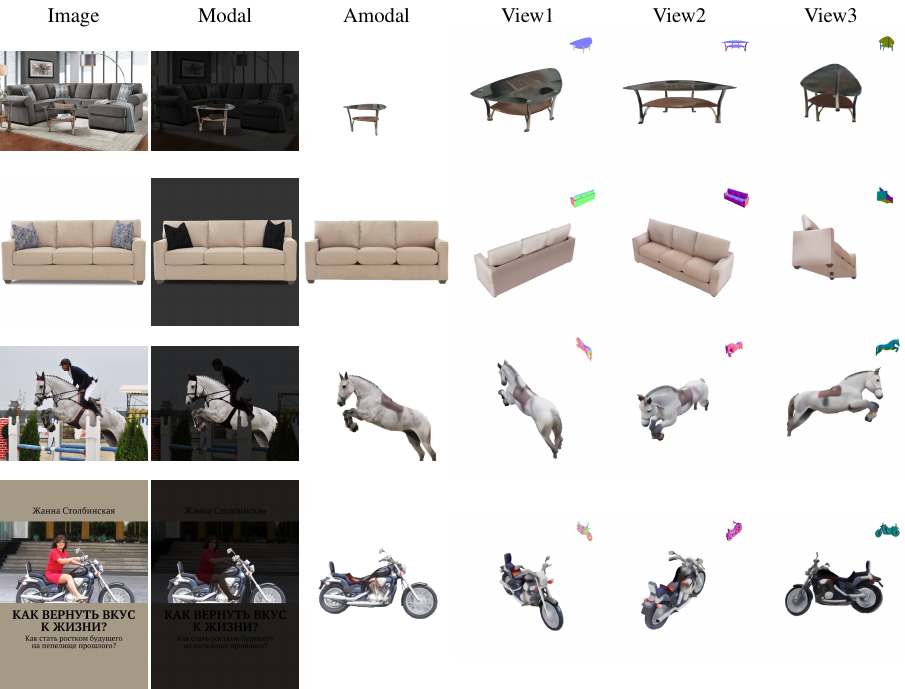}
    \caption{
        \textbf{Results on amodal 3D reconstruction.}
    }
    \label{fig.3d}
\end{figure*}

Our model can serve as a plug-and-play module to recover complete 3D geometry from single-view observations of occluded objects when combined with 3D generative models~\cite{liu2023zero, liu2023one, xu2023dmv3d, xu2024instantmesh}.
We demonstrate the capability of our model in amodal 3D reconstruction by integrating the deocclusion results into a state-of-the-art image-to-3D model, TRELLIS~\cite{xiang2024structured}, as shown in Fig.~\ref{fig.3d}.
The input images are from LAION~\cite{schuhmann2022laion} dataset, which are unseen during our deocclusion training.
The results highlight that our model can recover amodal RGBA with high compatibility with TRELLIS, enabling 3D AIGC products to handle occlusions more effectively in real-world scenarios, where occlusions are highly prevalent.
%

\section{Statistics and Examples of SynergyAmodal16K}

\begin{figure*}[!t]
    \centering
    \includegraphics[width=1\linewidth]{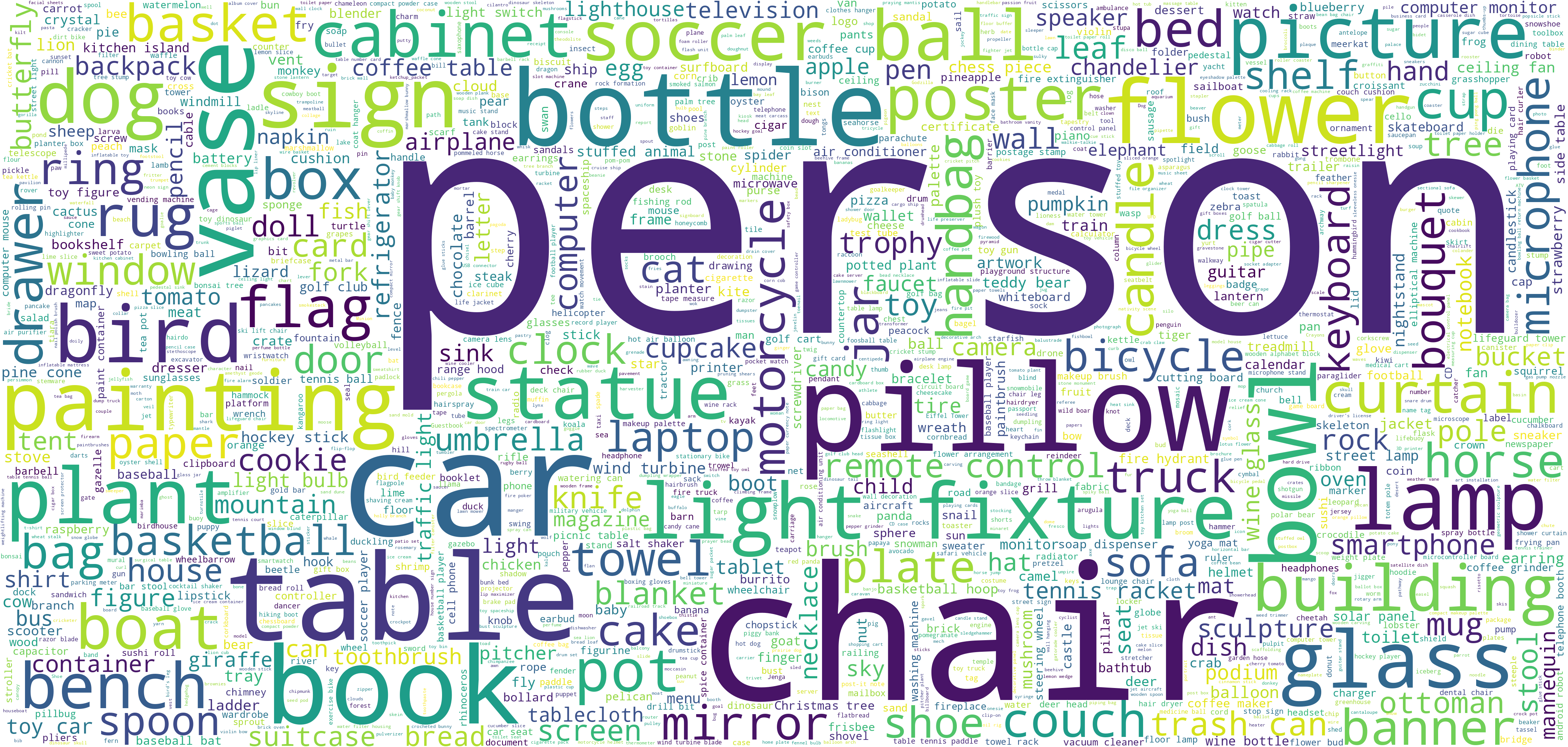}
    \caption{
        \textbf{Word Cloud.}
    }
    \label{fig.wordcloud}
\end{figure*}

\begin{figure*}[!t]
    \centering
    \begin{subfigure}[t]{0.33\textwidth}
        \centering
        \includegraphics[width=\linewidth]{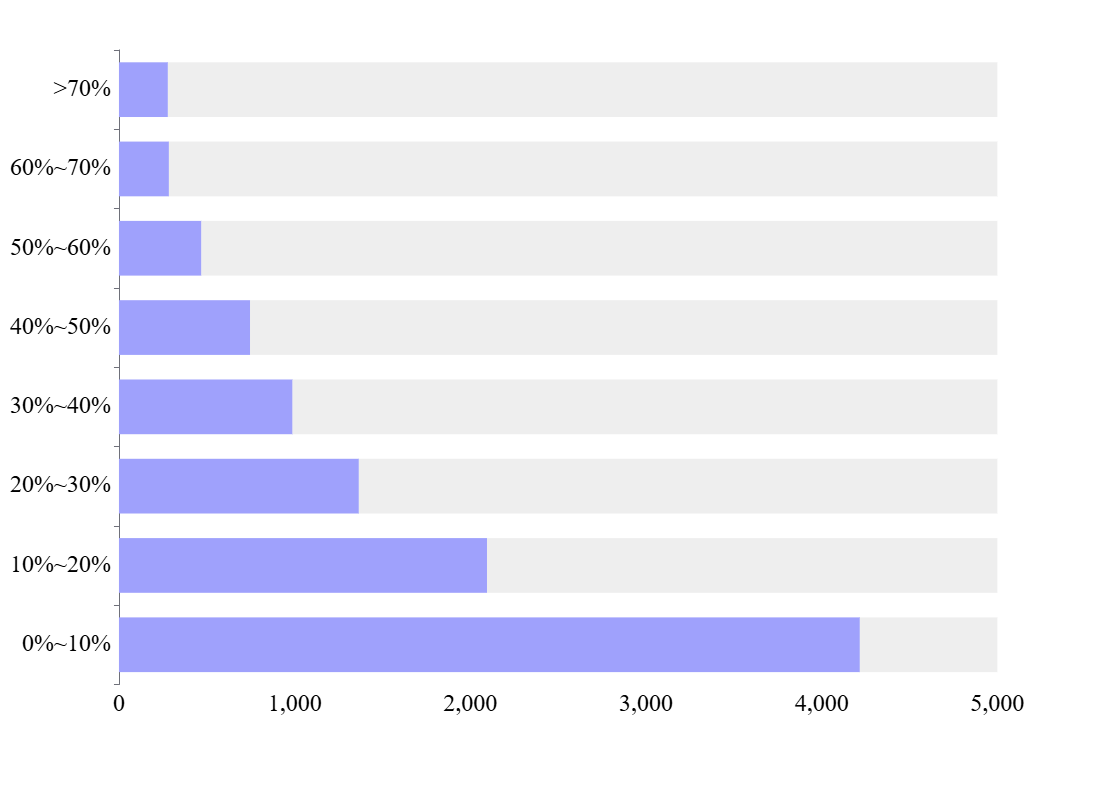}
        \caption{\textbf{Occlusion percentage}}
        \label{fig.chart1}
    \end{subfigure}
    \hfill
    \begin{subfigure}[t]{0.33\textwidth}
        \centering
        \includegraphics[width=\linewidth]{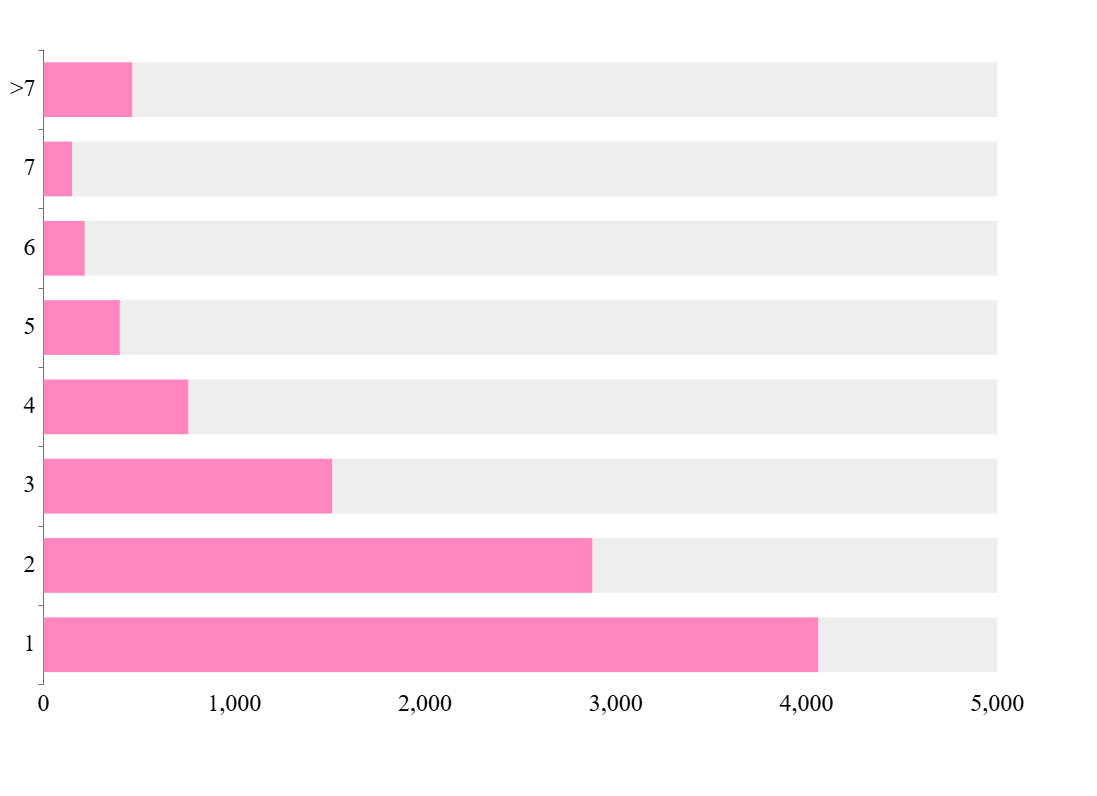}
        \caption{\textbf{Occluder number}}
        \label{fig.chart2}
    \end{subfigure}
    \hfill
    \begin{subfigure}[t]{0.33\textwidth}
        \centering
        \includegraphics[width=\linewidth]{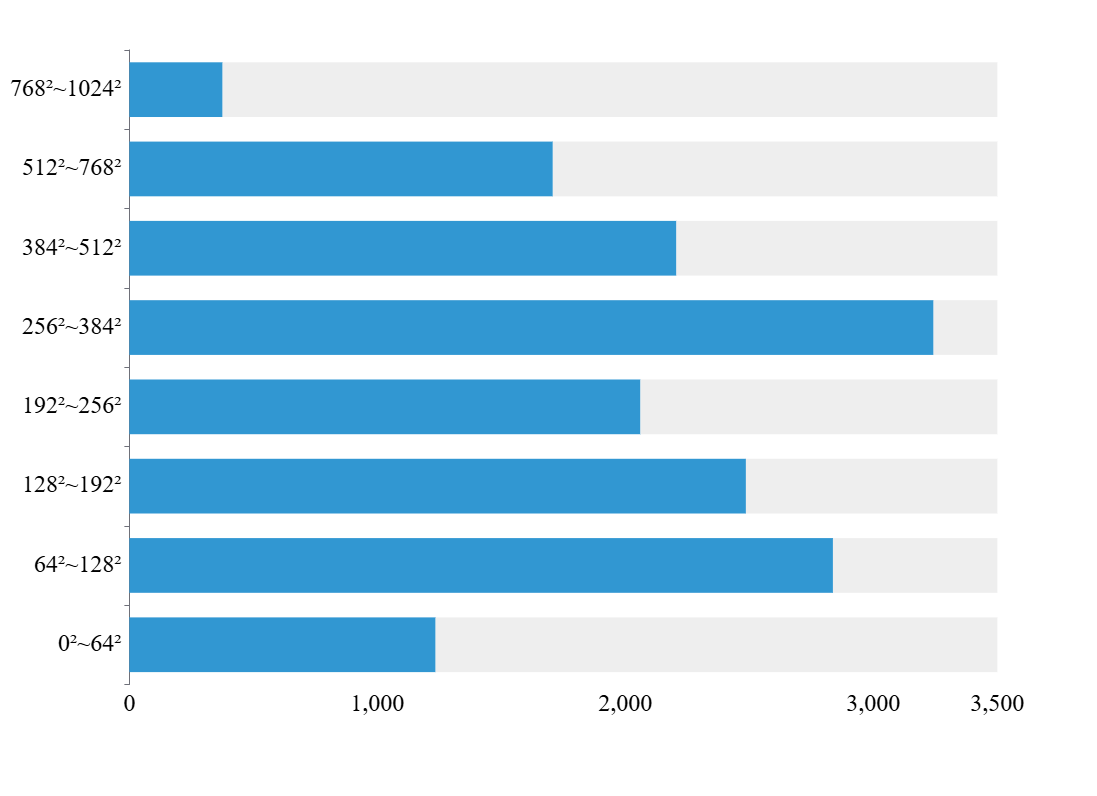}
        \caption{\textbf{Amodal resolution}}
        \label{fig.chart3}
    \end{subfigure}
    \caption{Statistics of SynergyAmodal16K dataset.}
    \label{fig.allcharts}
\end{figure*}

\begin{figure*}[!t]
    \centering
    \includegraphics[width=1\linewidth]{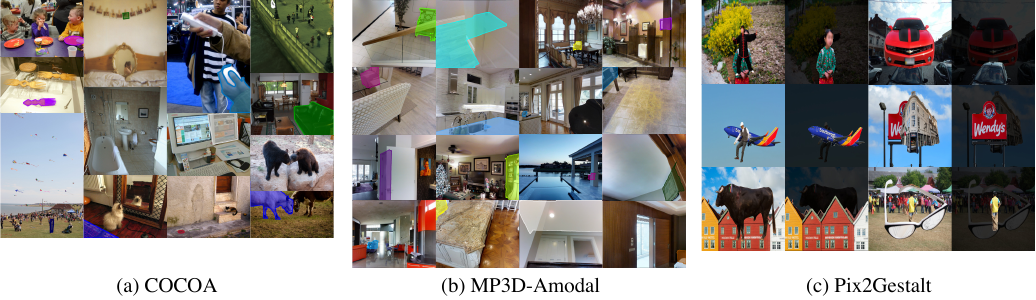}
    \caption{
        \textbf{Examples of other existing amodal datasets.}
    }
    \label{fig.others}
\end{figure*}

\begin{figure*}[!t]
    \centering
    \includegraphics[width=0.78\linewidth]{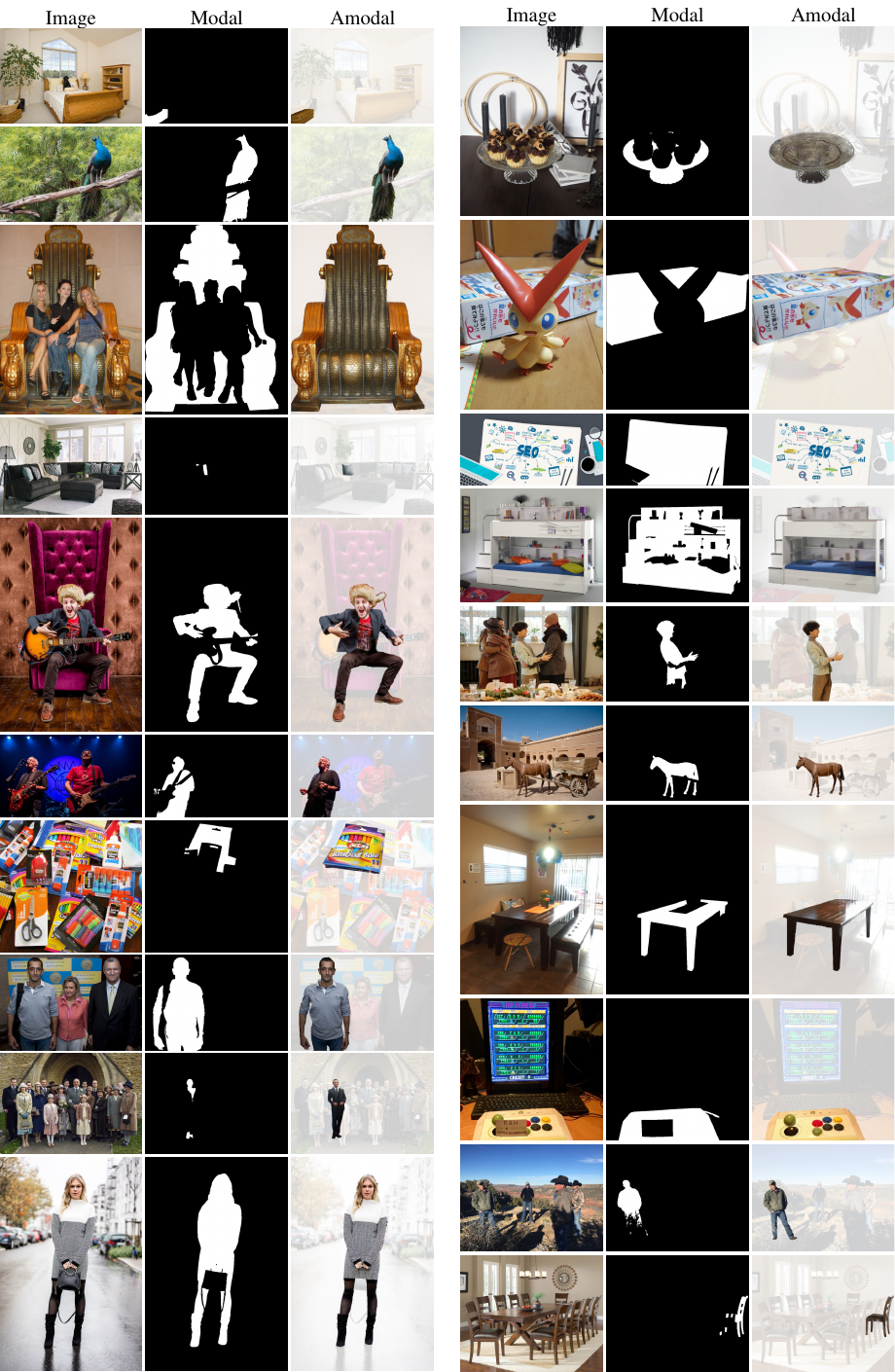}
    \caption{
        \textbf{Examples of SynergyAmodal16K dataset.}
    }
    \label{fig.dataset}
\end{figure*}

We additionally present more unfiltered examples from our proposed SynergyAmodal16K dataset in Fig.~\ref{fig.dataset}. As shown, our dataset encompasses a wide variety of scenes and objects.
We also display a word cloud of amodal object categories in Fig.~\ref{fig.wordcloud}, where the frequency of each category is represented by the size of the text. 
It can be observed that our dataset covers most common object categories encountered in daily life.
Furthermore, we include statistical graphs in Fig.~\ref{fig.allcharts} that detail the distributions of occlusion percentage, occluder number and amodal resolution. 
These statistics demonstrate that our dataset includes diverse occlusion scenarios and object sizes, thereby aiding the generalization of our method to general occlusion tasks.
We also present visualizations of existing amodal datasets for comparison in Fig.~\ref{fig.others}. It can be observed that:
COCOA~\cite{zhu2017semantic} demonstrates fair diversity but lacks appearance annotations;
MP3D-Amodal~\cite{zhan2024amodal} is limited to indoor objects and lacks diversity in general;
The dataset proposed by Pix2Gestalt~\cite{ozguroglu2024pix2gestalt}, based on SA-1B~\cite{kirillov2023segment}, lacks physical plausibility and does not cover diverse occlusion scenarios.
In comparison, our dataset stands out with the most detailed annotations and the most diverse and realistic occlusion scenarios.

\section{More Qualitative Comparison}

\begin{figure*}[!t]
    \centering
    \includegraphics[width=0.9\linewidth]{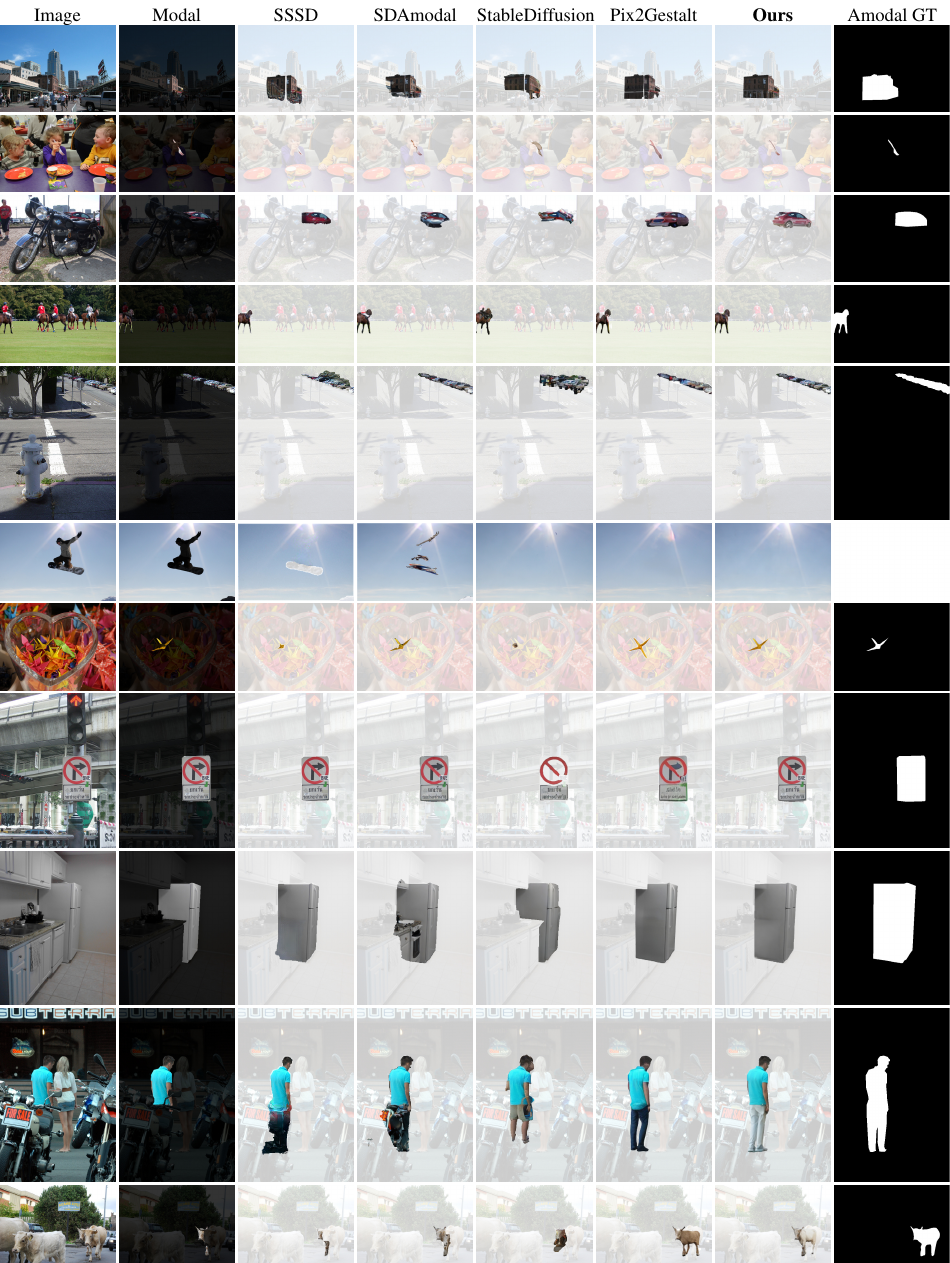}
    \caption{
        \textbf{More qualitative comparison.}
    }
    \label{fig.qualitative1}
\end{figure*}

\begin{figure*}[!t]
    \centering
    \includegraphics[width=0.9\linewidth]{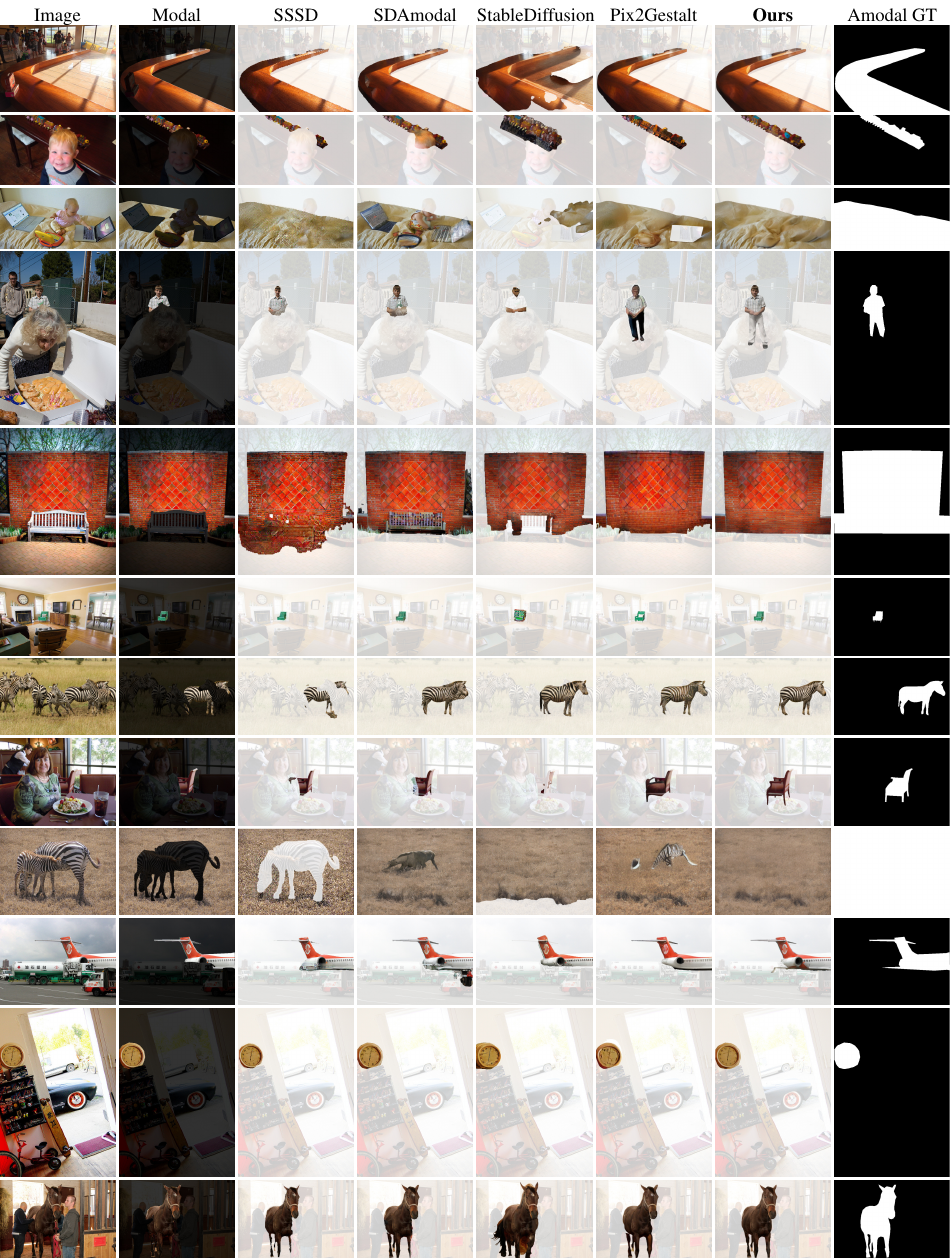}
    \caption{
        \textbf{More qualitative comparison.}
    }
    \label{fig.qualitative2}
\end{figure*}

We present additional qualitative comparison experiments in Fig.~\ref{fig.qualitative1} and Fig.~\ref{fig.qualitative2}. 
The results shown here are not cherry-picked.
It can be observed that our model achieves relatively better amodal shapes and appearances compared to all competing methods, with stable performance across various scenarios. This further demonstrates the effectiveness and generalization capability of our method.

\clearpage

\end{document}